\def\year{2022}\relax
\documentclass[letterpaper]{article} 
\usepackage{aaai22}  
\usepackage{times}  
\usepackage{helvet}  
\usepackage{courier}  
\usepackage[hyphens]{url}  
\usepackage{graphicx} 
\usepackage{natbib}  
\usepackage{caption} 
\DeclareCaptionStyle{ruled}{labelfont=normalfont,labelsep=colon,strut=off} 
\usepackage{algorithm}
\usepackage{algorithmic}

\usepackage{newfloat}
\usepackage{listings}
\floatstyle{ruled}
\newfloat{listing}{tb}{lst}{}
\floatname{listing}{Listing}
\title{Transferability of HRI Research: Potential and Challenges}
\author{
    Wafa Johal
}
\affiliations{
    
    University of Melbourne, Australia\\
    wafa.johal@unimelb.edu.au
}

\usepackage{bibentry}

\begin{document}

\maketitle

\begin{abstract}
With advancement of robotics and artificial intelligence, applications for robotics are flourishing. Human-robot interaction (HRI) is an important area of robotics as it allows robots to work closer to humans (with them or for them). 
One crucial factor for the success of HRI research is transferability, which refers to the ability of research outputs to be adopted by industry and provide benefits to society. 
In this paper, we explore the potentials and challenges of transferability in HRI research.
Firstly, we examine the current state of HRI research and identify various types of contributions that could lead to successful outcomes. Secondly, we discuss the potential benefits for each type of contribution and identify factors that could facilitate industry adoption of HRI research.
However, we also recognize that there are several challenges associated with transferability, such as the diversity of well-defined job/skill-sets required from HRI practitioners, the lack of industry-led research, and the lack of standardization in HRI research methods. We discuss these challenges and propose potential solutions to bridge the gap between industry expectations and academic research in HRI.
\end{abstract}

\section{Introduction}
Assessing the impact of research innovation is crucial for determining its economic, social, environmental, and cultural contributions.  The Australian Research Council (ARC) provides a definition of research impact that states: "\textit{Research impact is the contribution that research makes to the economy, society, environment or culture, beyond the contribution to academic research.}" \cite{ARC}.
This definition distinguishes between the \textbf{outputs} of the research (such as knowledge advancement disseminated via publications), the \textbf{outcomes} (such as commercial products, job creation, spin-offs, or integration into policy), and the \textbf{benefits} (such as economic, quality of life, or workforce benefits). 
When focusing on the economic and industrial impact, the term \emph{technology transfer} is often used to describe the process of transforming research outputs into outcomes and benefits. 

\begin{figure}[h]
    \centering
    \includegraphics[width=\linewidth]{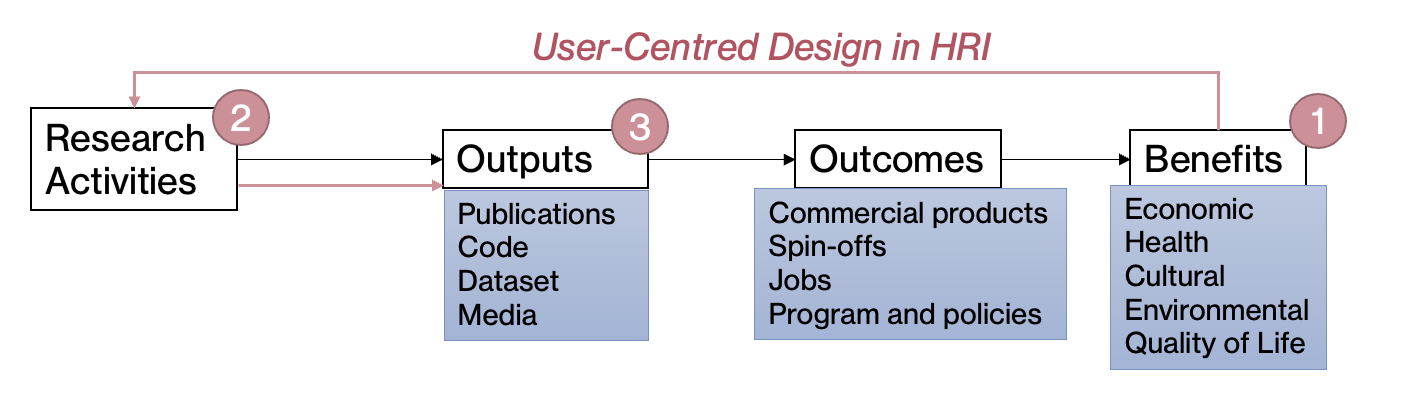}
    \caption{Typical research impact pathway in HRI, where the research will be initiated by a potential impact (1), but it will end with research outputs (3) without bridging back to outcomes.}
    \label{fig:pipleine}
\end{figure}

Researchers in Human-Robot Interaction (HRI) have often used user-centred design (UCD), participatory design or co-design, which are valuable methods to make sure a project will be beneficial to end-users \cite{Lupetti}. These design approaches focuses on the needs and preferences of the users throughout the design process, from the initial concept to the final product.
But while involving the target users, these methods often misses the bridge to an outcome that could allow the integration of the research into commercial product.
In other words, these design methods primarily focused on meeting the needs and preferences of the users throughout the design process, but they may not always consider the technical and logistical requirements necessary for the project to be integrated into an industrial pipeline. This can lead to a disconnect between the research and the practical applications of the technology.
Figure \ref{fig:pipleine}) shows an example of research pathway in a project using UCD.  Once the outputs are done, it is often hard for researchers to explore the potential outcomes as it may require the feasibility of the solution developed (is there a market? what is the added value of the technology?) 

In this paper, we explore the potential transfer of value from Human-Robot Interaction (HRI) research to industry by examining the types of research contributions generated by HRI research.

\section{A Short Overview of the types of HRI Research Contributions}
\begin{figure}
    \centering
    \includegraphics[width=\linewidth]{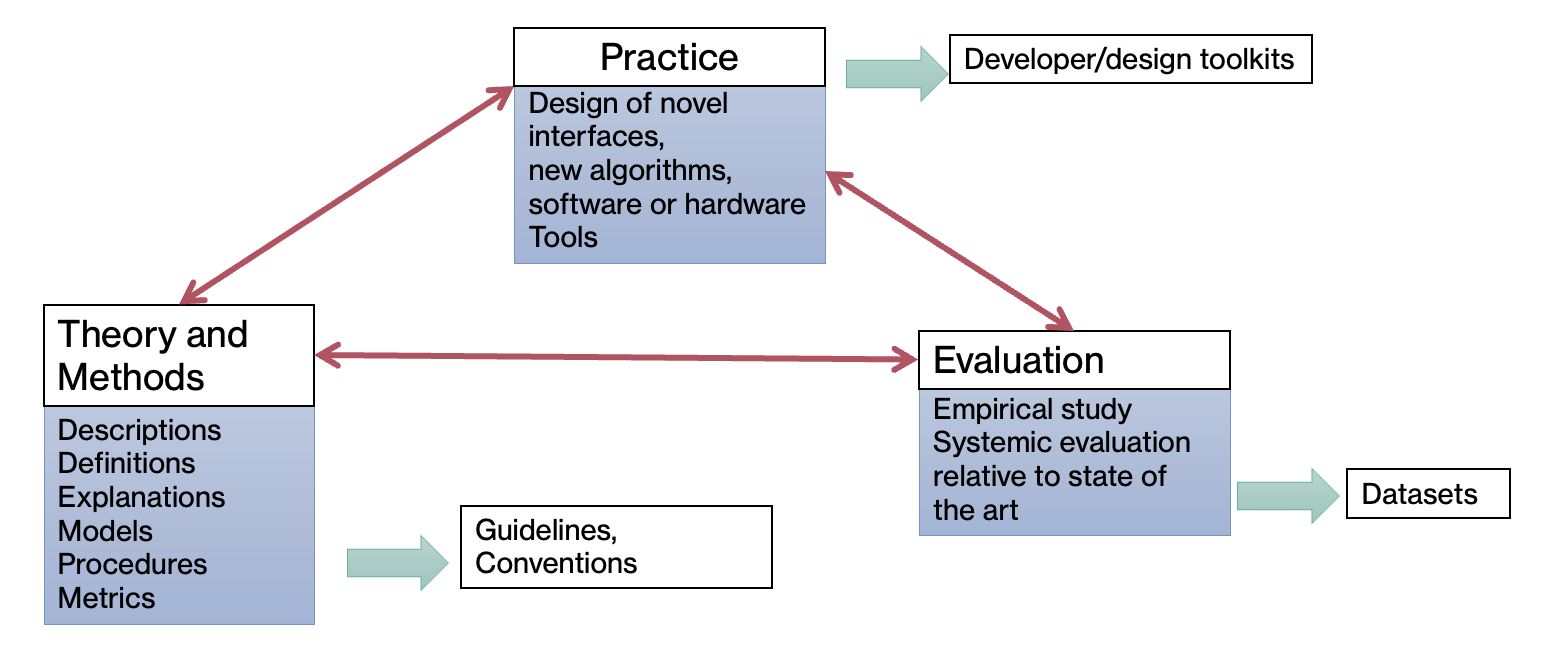}
    \caption{Types of Contributions in HRI- derived from \cite{SReeves}}
    \label{fig:tri}
\end{figure}

The field of Human-Robot Interaction (HRI) is highly interdisciplinary, encompassing a broad range of disciplines, from psychology and sociology to mechanical and computer sciences.
The ACM/IEEE Conference in Human-Robot Interaction \footnote{\url{https://humanrobotinteraction.org/category/conference/}} is one of the main conferences of the field and started in 2006. As the research community grew, the conference segmented the types of papers into themes. The current model of the conference now features 5 main themes: 1) User Studies, 2) Design, 3) Theory and Method and 4) Technical and 5) Systems. 
While these themes were designed for paper submissions, we propose to look at the types of contributions using a framework previously applied to HCI \cite{SReeves}. It distinguishes 3 types of contributions (Figure \ref{fig:tri}):
1) \textbf{Practice}: proposing the design of a novel interface, new algorithms, new software architecture or new hardware (usually submitted under Design and Technical/system track); 2) \textbf{Evaluation}: proposing the empirical study to understand the experience of the user interacting with the robot(s) - it could be qualitative or quantitative (usually submitted under the user-study track). and 3) the \textbf{Theory and Methods}: proposing formal descriptions, models, definitions, procedures or metrics. 

\section{Transferability Potentials}
Based on these different types of contributions, we derive potential outcomes that could be used for professional HRI practitioners.  

\paragraph{Practice:} Practice types of contributions can potentially be directly patented and turned into a product as beside the publication outputs, these types of research often generate a physical or software artefact (e.g. a library, a SDK). 
A key point to turn in into a product is to mobilise enough resources and funds to make the solution accessible to the industry.

\paragraph{Evaluation:} Empirical studies usually allow to verify a design or a model that was built. It also allows to loop back and generate theories or validate methods. Apart from the publications, studies also generate datasets, and in a world moving towards more data-greedy methods such as machine learning, data have a huge value. Unfortunately, most user studies presented at the HRI conference do not offer an open version of the data collected. 

\paragraph{Theory and Methods:} This type of contributions constitute the pedestal of scientifically grounded conventions and guidelines. For a field to establish models, professional procedures, measuring tools and formalised best practices allows professional practitioners to speed-up their design process and achieve robust solutions. 
For example, HCI designers and developers have worked on UI design principles and tools (e.g. color, spacing between visual elements, sizes ...) to assist practitioners (an instance of that is Material Design developed by Google \footnote{https://m3.material.io/}  which follow these conventions issue from research).

\section{Transferability Challenges}
There are still some challenges for HRI research to transfer to practice and commercialisation. We outlay a couple here, hoping that some investigations will be done to towards tackling them. 
\paragraph{Training for the job market.} Over the past decade, the number of Human-Robot Interaction (HRI) courses offered to students within higher degree programs has increased significantly. However, based on their syllabus, many of these courses focus heavily on research and theory rather than providing practical, hands-on experience. How can we develop hands-on curriculum that will train HRI practitioners? What skills are expected from the industry and how to design the learning objectives of HRI courses based on those? 
Becoming an expert in Human-Robot Interaction (HRI) requires not only knowledge but also practical experience \cite{lynham2002general}. While academia and research provide PhD students with a strong theoretical foundation, opportunities for designing and deploying HRI systems in real-world settings with target users are often lacking.
This lack of practical experience can make it difficult for students to demonstrate their expertise in the field. Without the opportunity to work on real-world projects, students may struggle to understand the practical challenges and considerations involved in designing effective HRI systems.

To address this challenge, it is essential to provide students with opportunities to gain practical experience in the field of HRI. This can include internships, industry collaborations, and hands-on projects that allow students to work on real-world problems. 
While these kinds of internships are often managed by universities, the HRI community could put in place some channels of communication to allow research students to find industry internship. 

\paragraph{Lack of Industry-led research.} When looking at the author's affiliation in the HRI conference, it is striking to see that there are extremely few papers for which the first authors is from the industry. Even more rare, are papers with authors only from the industry. 
On the other hand, when looking at CHI \cite{shneiderman_growth_2017}, we see that innovations that had the larger economical impact were often initiated by research conducted in the industry. 
How can the community facilitate and encourage industry led research? 

To encourage industry-led research in the field of HRI, conferences and other academics might need to be designed to better accommodate industry researcher. Designing opportunities for them to present their work, including them in the organisation of workshops and tutorials could also allow more collaboration between industry and academia. 

\paragraph{Lack of Norms and Standardisation in the field.}
As the field of HRI continues to grow, it is becoming increasingly important to develop theories and guidelines that can be integrated into professional practice. Similar to the HCI Fitt's Law which influenced the conventions in UI design, HRI should be able to produce models that are robust enough to be integrated into guidelines, conventions and processes in the design and implementation of HRI systems. But how? 
Shneiderman \cite{shneiderman_growth_2017} identifies two types of user studies in HCI: 1) the micro-HCI studies, focusing on well defined tasks and aiming to identify generic interaction principles (e.g. psychometric, ergonomics ...) and 2) the macro-HCI, which aims in studying the interaction in richer, more real-world scenarios. 
While researchers have been recently pushing for real-world experiments with target users, we also see that the theories and methods generated by the field are few; and this is due to the lack of micro-HRI. 
We argue that developing more rigorous ways of studying HRI will allow the development of standards that can be transferred more easily into general practice. Low hanging fruits in the domain would be to study safety issues and propose regulations and norms. This could be done by studying the perception of motion, varying the DoF for example.

\section{Conclusion}
As the societal distance between robots and end-users is reducing, it is important that the HRI field grows and generates productizable solutions. 
In order to facilitate this transfer, it would be good if researchers in the field would work on developing research transfer channels for each of their research project to easily highlight the commercial potential of their research outcomes. We believe that standardisation and open-science can good pathways toward more transfer between HRI research and the industry. 
In the future, we are thinking of developing a translational model similar to the one developed by \cite{10.1145/3290605.3300231} for HRI to facilitate the bridge between research outputs, outcomes and benefits.
As part of this project, we want to survey HRI professionals (in the industry) to understand their needs; and compare them to transfer strategies that academics in HRI have been using. We expect from this analysis to find and propose new ways to align the practitioners expectations with the outcomes generated by the research field.

\bibliography{references} 

\begin{thebibliography}{6}
\providecommand{\natexlab}[1]{#1}

\bibitem[{ARC(2023))}]{ARC}
ARC. 2023).
\newblock Research {Impact} {Principles} and {Framework} {\textbar}
  {Australian} {Research} {Council}.

\bibitem[{Colusso et~al.(2019)Colusso, Jones, Munson, and
  Hsieh}]{10.1145/3290605.3300231}
Colusso, L.; Jones, R.; Munson, S.~A.; and Hsieh, G. 2019.
\newblock A Translational Science Model for HCI.
\newblock In \emph{Proceedings of the 2019 CHI Conference on Human Factors in
  Computing Systems}, CHI '19, 1–13. New York, NY, USA: Association for
  Computing Machinery.
\newblock ISBN 9781450359702.

\bibitem[{Lupetti, Zaga, and Cila(2021)}]{Lupetti}
Lupetti, M.~L.; Zaga, C.; and Cila, N. 2021.
\newblock Designerly Ways of Knowing in HRI: Broadening the Scope of
  Design-Oriented HRI Through the Concept of Intermediate-Level Knowledge.
\newblock In \emph{Proceedings of the 2021 ACM/IEEE International Conference on
  Human-Robot Interaction}, HRI '21, 389–398. New York, NY, USA: Association
  for Computing Machinery.
\newblock ISBN 9781450382892.

\bibitem[{Lynham(2002)}]{lynham2002general}
Lynham, S.~A. 2002.
\newblock The general method of theory-building research in applied
  disciplines.
\newblock \emph{Advances in developing human resources}, 4(3): 221--241.

\bibitem[{Reeves(2014)}]{SReeves}
Reeves, S. 2014.
\newblock What {Is} the {Relationship} {Between} {HCI} {Research} and {UX}
  {Practice}? :: {UXmatters}.

\bibitem[{Shneiderman(2017)}]{shneiderman_growth_2017}
Shneiderman, B. 2017.
\newblock The {Growth} of {HCI} and {User} {Interface}/{Experience} {Design}:.

\end{thebibliography}

\section{Acknowledgments}
This research is supported by the Australian Research
Council Discovery Early Career Research Award (Grant No. DE210100858).

\end{document}